\documentclass[10pt,twocolumn,letterpaper]{article}

\usepackage{iccv}
\usepackage{times}
\usepackage{epsfig}
\usepackage{graphicx}
\usepackage{amsmath}
\usepackage{amssymb}
\usepackage{dirtytalk}
\usepackage{subcaption}
\usepackage[T1]{fontenc}

\iccvfinalcopy 
\def\iccvPaperID{****} 

\ificcvfinal\fi

\begin{document}

\title{Learning to Inpaint by Progressively Growing the Mask Regions}

\author{Mohamed Abbas Hedjazi, Yakup Gen\c{c}\\
Gebze Technical University\\
Kocaeli, Turkey\\
{\tt\small {mahedjazi,yakup.genc}@gtu.edu.tr}
}

\maketitle

\begin{abstract}
Image inpainting is one of the most challenging tasks in computer vision. Recently, generative-based image inpainting methods have been shown to produce visually plausible images. However, they still have difficulties to generate the correct structures and colors as the masked region grows large. This drawback is due to the training stability issue of the generative models. This work introduces a new curriculum-style training approach in the context of image inpainting. The proposed method increases the masked region size progressively in training time, during test time the user gives variable size and multiple holes at arbitrary locations. Incorporating such an approach in GANs may stabilize the training and provides better color consistencies and captures object continuities. We validate our approach on the MSCOCO and CelebA datasets. We report qualitative and quantitative comparisons of our training approach in different models.
\end{abstract}

\section{Introduction}
\label{section:introduction}

Image inpainting is a technique that allows filling in missing regions/holes or removing unwanted objects/artifacts in an image. This task is easy for humans since they can understand the image structure representing the scene, even when a significant portion of the scene is not visible. However, it is a very challenging task for a computer. It is applied in many problems including localization and segmentation \cite{1}, video compression \cite{2}, 3D shape inpainting \cite{3}, depth inpainting \cite{4}\cite{5} and face verification \cite{6}. 

Recently, deep learning methods \cite{17,18,19,20} applied Generative Adversarial Networks (GANs) \cite{21} to fill in masked regions by learning from large image datasets. They outperform the traditional inpainting methods \cite{14, 15,16} both qualitatively and quantitatively. However, some of these methods \cite{17} fill in the center of the image, that may fail to inpaint variable size regions. Furthermore, they suffer from artifacts around the inpainted regions and need post-processing steps to correct the resulted image \cite{18}. Therefore, understanding the structure and different objects in the scene helps to achieve high-quality image completion. 

Although GANs fit the inpainting problem very well, they suffer from stability problems that lead to mode collapse and over-fitting. To address these limitations, \cite{27} provides architectural guidelines and optimization hyper-parameters that leads to better synthesis results. Moreover, a multi-stage generation approach introduced in \cite{50} creates high-quality images by progressively adding layers to the generator and the discriminator. Furthermore, \cite{51} improves \cite{50} by controlling the visual features of the image in different scales through the adaptive normalization layer \cite{52}. Some works addressed the loss functions improving the training stability including Wasserstein distance \cite{29}, Least Squares \cite{29} and Energy-based GANs \cite{49}. 

Another attempt to stabilize the training of GANs is to employ a curriculum learning (CL) approach \cite{36}. It achieved a lot of success in many tasks, including natural language processing \cite{46} \cite{47}, image recognition \cite{48} and generation \cite{45}. CL is a setting in which it gradually reveals training samples to the model from the easiest to the most difficult. Inspired by this idea, we propose a curriculum-style strategy to progressively train an effective generator by growing the size of the masked regions in the context of image inpainting. The intuition was that the generator and the discriminator networks solve the inpainting problem starting from simpler to much harder inpainting regions. By simpler, we mean small masked regions with basic structures that can be filled easily without the need for global object structures. On the other hand, harder means larger mask regions that need both local and global understanding of the scene.

We validate our approach using several models of different architectures and loss functions. The first one is our customized model that is trained using two networks: a deep residual convolutional generator \cite{22}, and a multi-scale discriminator that criticizes the quality and the relevance of the completed image in different scales. In the generator, we replace the vanilla convolutions with the gated convolutions introduced in \cite{20}. They proved that it is a good replacement for vanilla convolutions in the context of image inpainting. The other methods are two of the state-of-the-art models \cite{17} and \cite{20}. We conduct two experiments: fixed versus progressively growing masked regions on the previously stated models. Additionally, to show the effectiveness of our approach, we check if a simple reconstruction loss is sufficient to stabilize the generator for the first training iterations. In another setup, we use a fixed masked region then gradually increase the adversarial loss weight. We report qualitative and quantitative results on the MSCOCO \cite{35} and CelebA \cite{23} datasets. The quantitative metrics include L1, PSNR, Inception score (IS) \cite{32} and Frichet Inception Distance (FID) \cite{54} quality metrics

Our contributions are as follow:
\begin{itemize}
\item We propose the progressively growing of the masked regions as a GAN stabilization technique for image inpainting tasks.

\item We compare the usage of fixed versus progressively growing mask regions using different architectures and loss functions, and we report the qualitative and quantitative results on two challenging datasets.

\item We investigate other training stabilization setups and compare it against our approach.
\end{itemize}

\section{Related Work} 
\label{section:relwork}

\begin{figure*}[h]
  \centering
  \includegraphics[keepaspectratio, width=0.9\textwidth]{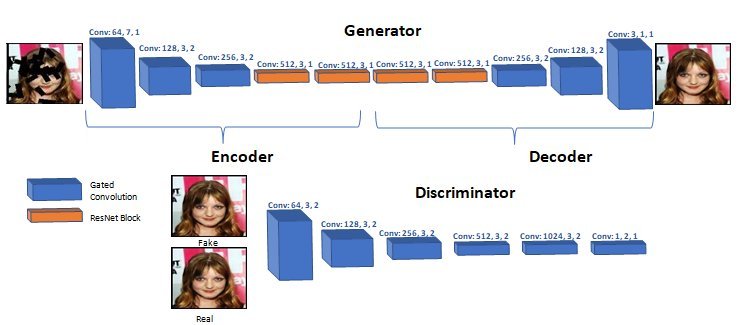}
  \caption{Illustration of the Generator (top) and the Discriminator (down) network architectures. The generator takes as input the masked image and the binary mask. It outputs the inpainted image. The discriminator takes either the ground truth (real) or the generated (fake) image as input and outputs either fake or real tensors.}
  \label{fig1}
\end{figure*}

\textbf{Traditional Inpainting}  methods such as the diffusion-based image synthesis propagates the closest pixels around the masked regions to fill it in \cite{7,9}. Nevertheless, these methods have many limitations because they just complete texture patterns and do not understand the anatomy of the scene and the objects to be completed. Moreover, they cannot fill in large masked regions. On the other hand, patch-based methods can fill in large masked regions in images by searching for similar patches in the image \cite{14,15,16}. However, these methods fail to fill in large holes in complex scenes especially when the texture to be filled is not present in the image. Furthermore, patch-based methods are slow, come at a large processing cost and are not based on a semantic understanding of the scene.  Therefore, the inpainting task cannot be handled by traditional inpainting approaches since the missing region is very large for local-non-semantic methods to work well.

\textbf{Deep Learning-based inpainting} methods fill in masked values in an end-to-end manner by optimizing a deep encoder-decoder network to reconstruct the input image. But, it tends to produce blurry images and often proceeded by a post-processing step. To outperform this limitation, GANs \cite{21} showed to be a great data distribution modeling technique synthesizing realistic-looking images. GANs train two networks against each other in a minimax game, the first one generates images from a random distribution (called generator) and the later which tries to distinguish between real and generated images (called discriminator).

In the context of image inpainting, \cite{17} optimizes an encoder-decoder network to produce relevant content in a central rectangular hole in the image based on GANs. This approach can generate novel objects and textures in the image, but it lacks local consistency. To address this limitation, \cite{18} extends it using a global and a local discriminator to ensure both the global coherence and the local image consistency. The drawback of this technique is the post-processing step that should be done using Poisson Image blending \cite{25}. 

Another valuable work is \cite{26} that presents a novel contextual attention layer to explicitly attend on related feature patches at distant spatial locations. \cite{19} uses a stack of partial convolution layers and mask updating steps to perform image inpainting using an autoencoder without adversarial learning. The intuition was that regular convolutions treat both valid pixel values and masked values in the same manner while partial convolutions are conditioned only on valid pixels. The proposed architecture demonstrates the effectiveness of training image inpainting models on irregularly shaped holes.

Moreover, \cite{20} introduces Gated Convolutions to overcome the limits of \cite{19}. The latter is a hard-gating single-channel un-learnable layer multiplied to input feature maps. However, Gated Convolutions are learnable layers that learn a dynamic feature selection mechanism for each channel and each spatial location. Additionally, it allows user-input (a sketch) as an additional channel.

\textbf{Curriculum Learning:} is an effective approach to improve the training of neural networks. Unlike the traditional training approach of CNNs that uniformly samples mini-batches from the data distribution, \cite{42} used CL to order the training samples by difficulty and creates mini-batches from them, this lets the network start with the easiest ones which improve both the accuracy and the learning speed. Further, \cite{42} improves the generalization ability by increasing the dropout rate throughout training that gradually increases the difficulty of the problem.
 
Furthermore, \cite{40} employed CL on GANs by making the discriminator solves harder problems during training. They augment the dimensionality of the sample space with additional random variables. This approach makes the task much difficult for the discriminator and prevents it from being over-confident.

In the context of image inpainting, \cite{45} utilizes a progressive generative network to fill-in images with squared masks. The approach splits the task into different stages, where each one aims to do a part of the entire curriculum. After that, an LSTM framework is used to chain all of them together. \cite{44} utilize CL on the contour and image completion modules using different stages. The training starts using only the content loss, then they fine-tune it with a small weight for the adversarial loss. Finally, they fine-tune the whole module using 1:1 weights.

Unlike the previously mentioned methods, we adapt CL in the image inpainting task by progressively growing the masked regions during training. In test time, the generator network can fill irregularly shaped holes. To the best of our knowledge, no method uses a similar approach in the context of image inpainting.

\section{Progressive Image Inpainting}\label{section:pii}

\begin{figure}
\centering
\includegraphics[width=0.45\textwidth]{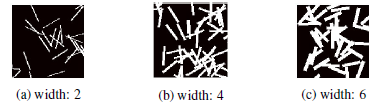}
\caption{Illustration of the progressively growing approach. After each k iterations, we progressively increase the masked region size until it reaches half the size of the image.}
\label{fig2}
\vspace{-3mm}
\end{figure}

As it is known, GANs are very hard to train due to its nature that depends on two networks having two sets of parameters optimized independently of each other. That leads to many problems, including mode collapse, non-convergence, and the vanishing gradients. The inpainting task is strongly affected by robust adversarial loss functions, stable architectures, and GAN stabilization techniques. We focus on the last point and propose a simple yet effective training technique to stabilize the training of GANs in the context of image inpainting. The process is as follows: the generator starts by solving a simple problem successfully, after each k iterations the masked region grows to a much harder problem till the region size reaches the half size of the image, as illustrated in Figure~\ref{fig2}. By simple, we mean that the masked region contains basic structures (textures) while hard refers to masked regions that contain complicated structures and objects.

We claim that, in the beginning, the generator easily fills-in the narrow masked region since the adversarial loss is responsible for an easy problem that is simply a reconstruction in this case. Then, the problem's difficulty increases as we grow the width of the mask. Consequently, the generator can fill-in the half size of the masked region without much difficulty. That makes the adversarial loss stable in the next k training iterations. The training process continues this way till a  specified maximum width. We will investigate this claim by reporting the quantitative results of each k iterations using different mask sizes.

\section{Architectures and Training} 
\label{section:at}

To validate our approach, we use different models: our customized model illustrated in Figure~\ref{fig1}, the Context-encoder model \cite{17} and the Free-form inpainting model \cite{20}.

\textbf{Our customized model:} the generator has two sub-networks, an encoder network that down-samples the size of the input to 1/4 the original size followed by two residual blocks. We duplicate the number of filters after each gated convolution and residual block. The decoder network is the reverse order of the encoder. Instead of using transposed convolutions as generally done in decoders, we use bilinear interpolation before applying gated convolutions. The last convolution layer outputs an RGB image. 

In the discriminator network, we use a multi-scale architecture that contains five convolution layers. It downsamples the feature maps size and increases the number of filters. The last two convolution layers have the same number of filters. The discriminator outputs an array of network layers on different scales.

Instead of using Batch Normalization \cite{30} that seems to cause problems in the inference time, especially when the batch size is small, we use the Instance Normalization \cite{31} that normalizes each batch independently across spatial locations. Additionally, it provides visual and appearance in-variance, moreover it is agnostic to the contrast of the image.  The loss functions include the LSGAN loss \cite{28}, an L1 loss between the non-masked regions in the ground truth and the generated image, an L1 loss between the masked region in the ground truth and the generated image, finally, we include the Perceptual loss using a pre-trained VGG network \cite{34}.

\textbf{The Context-encoder model:} optimizes an auto-encoder network to produce a rectangular hole in the center of the image. The discriminator considers the later as fake, while the center of the ground truth image as real.  The training requires two loss functions: a pixel-wise reconstruction loss and an adversarial loss \cite{21}.

\textbf{The Free-form inpainting model:} the generator has the same architecture as \cite{26} followed by a refinement network without residual connections. The discriminator is a PatchGAN that classifies image patches of size 70x70 as real or fake. Thus, there is no need for a global and local discriminator as in \cite{18}. Furthermore, the networks do not add any normalization layer. It computes two loss functions: the Hinge loss and the reconstruction loss. It does not include any perceptual or style loss.

\section{Experimental setups}
\label{section:ep}

In this section, we describe the datasets, the experimental setups, and the comparisons planning.

\textbf{Datasets:} we experiment on a variety of challenging datasets including MSCOCO \cite{35} and CelebA \cite{23}. The first dataset contains cluttered scenes with a lot of changes in colors and structures.  The later dataset contains cropped faces that have fewer structure changes. We train on 200k and 82k training images defined in CelebA and MSCOCO, respectively. We test the performance on 10000 random validation images (no available test set) for the CelebA dataset and 5000 test images for the MSCOCO dataset.

\textbf{Experimental setups:} our main experimental setup is to investigate the fixed size masks versus the progressive growing approach. We use constant weights for both the reconstruction and the adversarial loss. To prove/disprove our claim and hypothesis in~section \ref{section:pii}, we are planning to explore the following setups using a fixed mask size:

\begin{itemize}
\item Use a simple reconstruction loss for the first k iterations, then include the adversarial loss where both loss functions will have fixed weights.
\item Fix the reconstruction loss during the whole training and increase the adversarial loss weight after each k iterations.
\end{itemize}

\textbf{Comparison plan:} unlike the common comparison showing the outperformance of their method against the SOTA, in our case, we aim to confirm the impact of our proposed training scheme (Progressive growing) and the two other setups described above. We use different models including our customized model, CE \cite{17}, and Gated \cite{20}. To adapt our training approach to the CE model, we are planning to start the training process with a small rectangle in the middle, then progressively increase the rectangle size to reach the half size of the image.  \cite{20} adds the sketch as an additional input to the model. To ensure a fair comparison, we only input the image and the mask. 

We test on the MSCOCO and CelebA datasets for the different setups on our customized model and \cite{20}.  We report the quantitative comparison using L1, PSNR, IS, and FID. Furthermore, we show the output of our customized model versus \cite{17} on different training schemes in the qualitative comparison. Since the CE model input is a fixed central mask in the middle of the image, we do not compare it against the other models. Thus, we only report the qualitative and quantitative results of the different setups against each other. We do not perform any post-processing step for all the models. Due to hardware restrictions, we use images of resolution 128x128 in both datasets.\linebreak

We implement the models using Pytorch v1.1.0, CUDA v10.0, CUDNN v7.5.1, and the hardware GPU is NVIDIA GTX 1080 Ti. The training takes roughly five days per experiment.

\section{Implementation details} 
\label{sec:implementation}

\begin{figure}[h]
  \centering
  \begin{subfigure}[b]{0.5\textwidth}
     \centering \includegraphics{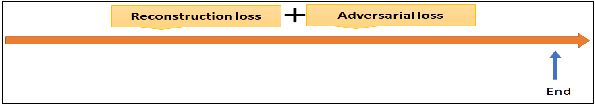}
     \caption{}
  \end{subfigure}
  
    \hspace{5pt}
  \begin{subfigure}[b]{0.5\textwidth}
     \centering \includegraphics{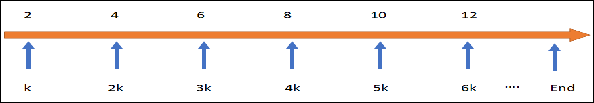}
     \caption{}
  \end{subfigure}
  
  \hspace{5pt}
  \begin{subfigure}[b]{0.5\textwidth}
     \centering \includegraphics{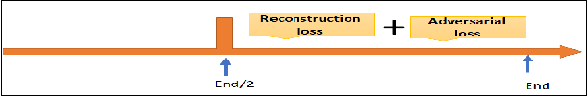}
     \caption{}
  \end{subfigure}
  
  \hspace{5pt}
  \begin{subfigure}[b]{0.5\textwidth}
     \centering \includegraphics{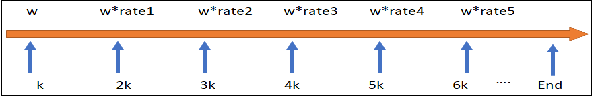}
     \caption{}
  \end{subfigure}
  
  \caption{Different training setups of our experiments. (a) uses the reconstruction loss and the adversarial loss for all the iterations. (b) is our progressive growing masks region approach. (c) uses only the reconstruction loss for the first half of training and adds the adversarial loss in the second half. (d) increases the adversarial loss weight after k iterations.
}
  \label{fig:setups}
\end{figure}

In this part, we show the impact of our proposed training approach on the customized and the state-of-the-art models \cite{17} \cite{20}. As mentioned previously, we compare the fixed versus progressive mask size approach for all the models on the MSCOCO and CelebA datasets. Furthermore, to prove/disprove the correctness of the proposed method, we compare it with two other training strategies, as shown in Figure~\ref{fig:setups}. For the customized model, both the generator and the discriminator networks use Adam as the optimizer with a learning rate of 0.0002 and a batch size of 4. For \cite{17} and \cite{20}, we keep the same hyper-parameters used in the original work. We train all the models for 1M iterations. Further, we increase the mask size and the adversarial weight after 100k iterations. For a fair comparison, we fix the randomness seed while training the models to ensure that we give the same input (same masked regions) and the same order of the images to the models.

\begin{figure*}[h]
  \centering
  \includegraphics[keepaspectratio, width=0.9\textwidth]{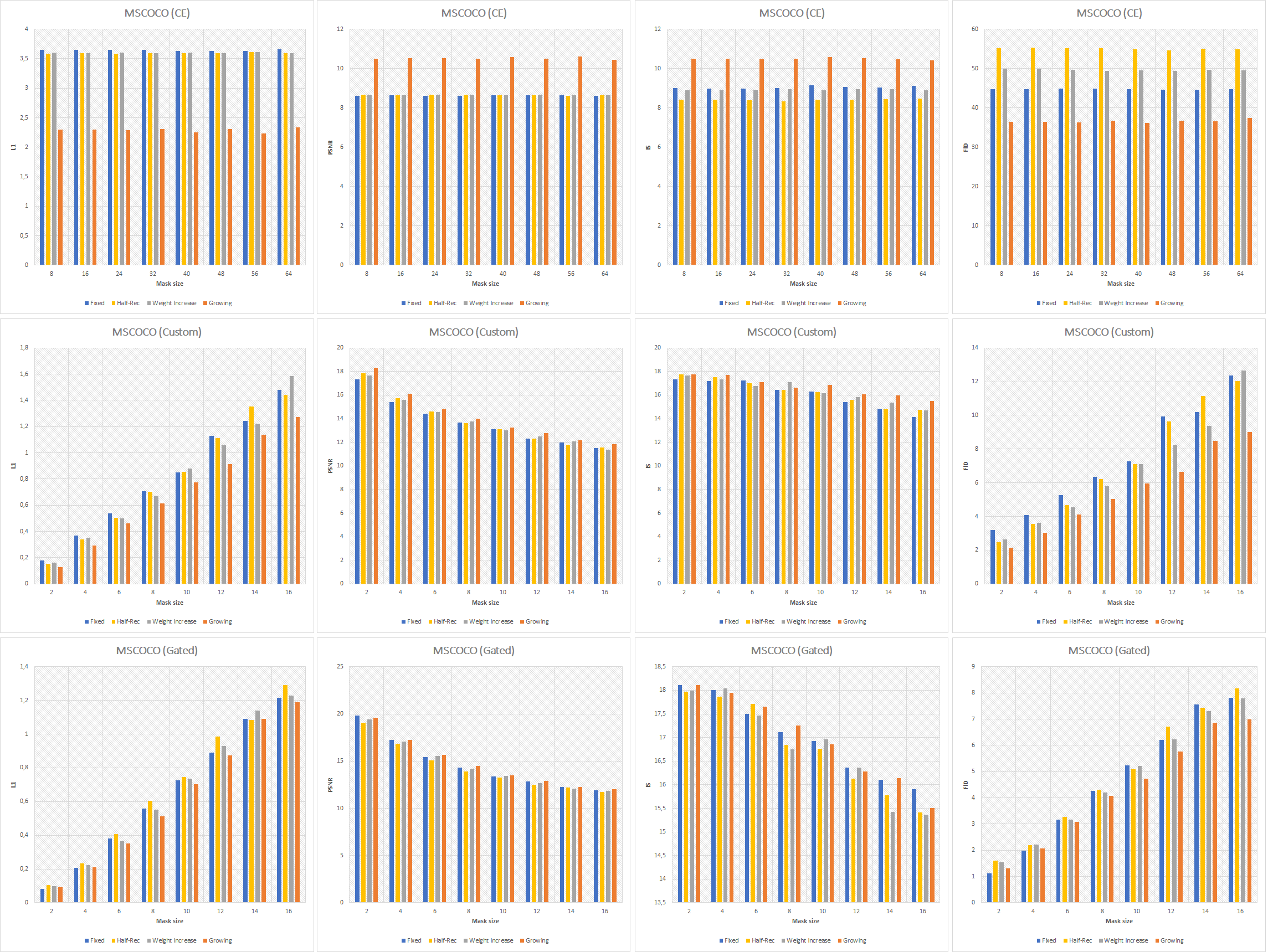}
  \caption{Quantitative comparison of the different training approaches on the MSCOCO datasets. Each column from left to right shows the L1, PSNR, IS, and FID scores. Each row from top to down shows: \cite{17}, our customized model and \cite{20}.}
  \label{quantify_mscoco}
\end{figure*}

\begin{figure*}[h]
  \centering
  \includegraphics[keepaspectratio, width=0.9\textwidth]{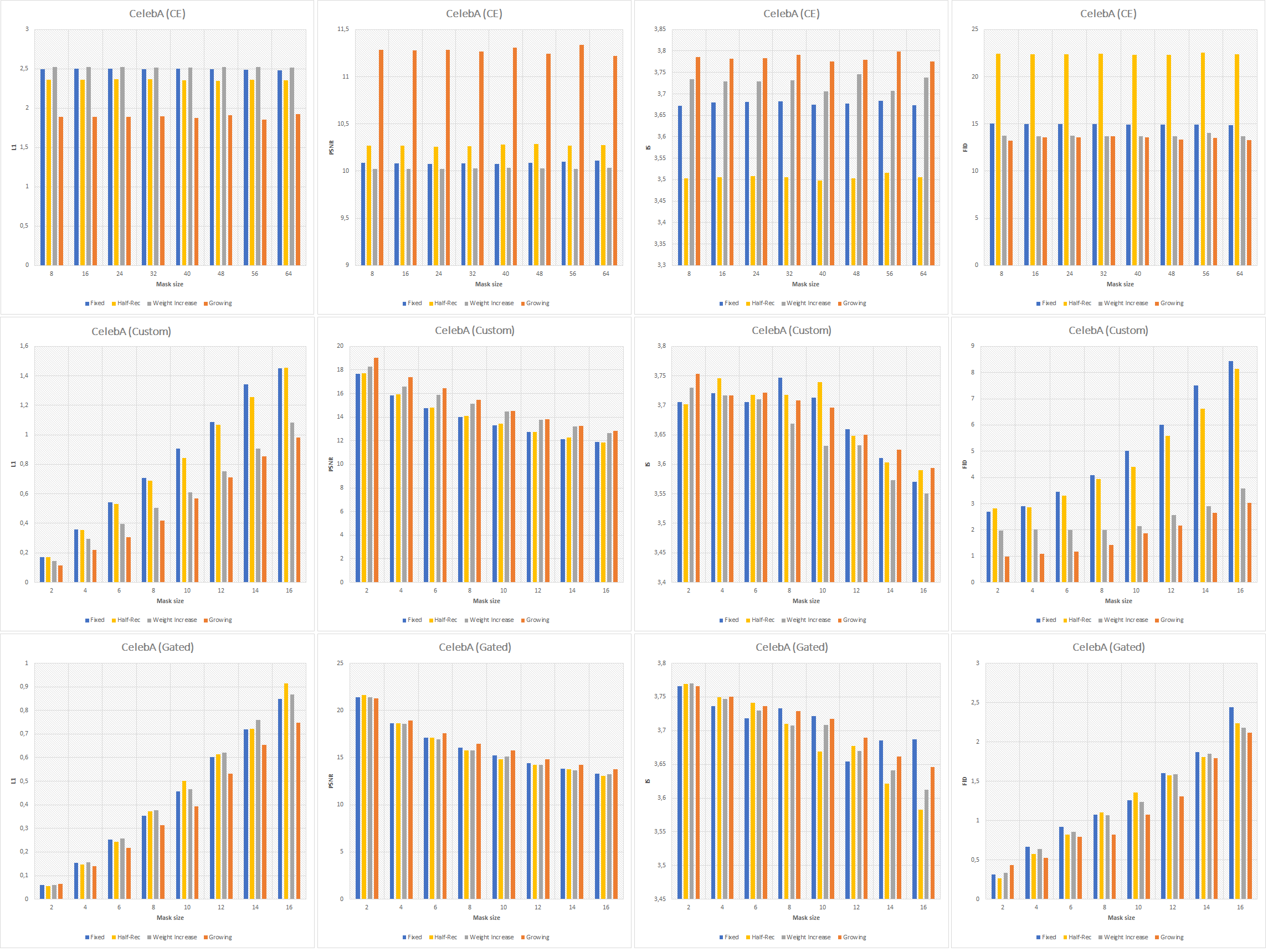}
  \caption{Quantitative comparison of the different training approaches on the CelebA datasets. Each column from left to right shows the L1, PSNR, IS, and FID scores. Each row from top to down shows: \cite{17}, our customized model and \cite{20}}
  \label{quantify_celeba}
\end{figure*}

\section{Experimental results}
\label{sec:results}

\subsection{Quantitative Evaluation}
\label{sec:quantitative}

We experiment using common evaluation metrics including L1, PSNR, IS and FID to quantify the performance of the training approaches. The L1 and PSNR are calculated using the model's outputs and the ground truth images. For the IS and FID scores, we use an InceptionV3 model \cite{53} pretrained on the ImageNet dataset \cite{24}. 

From Figure~\ref{quantify_mscoco} and Figure~\ref{quantify_celeba}, we see that our progressive growing approach improves the performance (L1, PSNR, and FID) of all the models in the MSCOCO and CelebA datasets. Meanwhile, the other three setups results are not deterministic, since they outperform each other depending on the model and the dataset. The IS is based on the classification probabilities, therefore it does not give a stable performance quantification on CelebA dataset since the latter one contains only one class (faces).

As claimed previously, to prove the effectiveness of our training approach, we experiment using different mask sizes. In most cases, our approach outperforms the other setups on the two datasets as shown in Figure~\ref{quantify_mscoco} and Figure~\ref{quantify_celeba}.

To apply our curriculum learning training approach to free-form mask models, we have to control the width, height, orientation, and the number of masks in the images. On the other hand, applying it on \cite{17} is easier since we can control the size of the masked regions (rectangular mask shapes). Nevertheless, the performance of this model is still low compared to the other models due to its local consistency nature and the use of the standard convolution layers. 

Although our customized model has a larger number of parameters than \cite{20}, the later outperforms it in all the training approaches in the MSCOCO and CelebA datasets as shown in ~Table \ref{tab:mscoco} and ~Table \ref{tab:celeba}. This can be explained by the usage of a refinement network in \cite{20}.

\begin{table}[h!]
\begin{center}
\begin{tabular}{|p{2.5cm}||p{1.5cm}|p{1.5cm}|}
\hline
Metric & Fixed & Growing\\
\hline
L1 (CE) & 3,6570 & \textbf{2,3368} \\
L1 (Custom) & 1,4814 & \textbf{1,2737} \\
L1 (Gated) & 1,2168 & \textbf{1,1891}\\
\hline
PSNR (CE) & 8,6192 & \textbf{10,4514}\\
PSNR (Custom) & 11,5228 & \textbf{11,8406}\\
PSNR (Gated) & 11,9331 & \textbf{12,0229}\\
\hline
IS (CE) & 9,1214 & \textbf{10,4155}\\
IS (Custom) & 14,1323 & \textbf{15,5197}\\
IS (Gated) & \textbf{15,8999} & 15,5010 \\
\hline
FID (CE) & 44,7089 & \textbf{37,3982}\\
FID (Custom) & 12,3523 & \textbf{9,0147}\\
FID (Gated) & 7,8109 & \textbf{6,9836}\\
\hline
\end{tabular}
\caption{Quantitative comparison between the fixed and the progresive growing mask using the state-of-the-art models and our custom model on the MSCOCO dataset.}
\label{tab:mscoco}
\end{center}
\end{table}

\begin{table}[h!]
\begin{center}
\begin{tabular}{|p{2.5cm}||p{1.5cm}|p{1.5cm}|}
\hline
Metric & Fixed & Growing\\
\hline
L1 (CE) & 2,4798 & \textbf{1,9233}\\
L1 (Custom) & 1,4500 & \textbf{0,9807}\\
L1 (Gated) & 0,8491 & \textbf{0,7479}\\
\hline
PSNR (CE) & 10,1082 & \textbf{11,2184}\\
PSNR (Custom) & 11,9022 & \textbf{12,8514}\\
PSNR (Gated) & 13,3147 & \textbf{13,7816}\\
\hline
IS (CE) & 3,6729 & \textbf{3,7752}\\
IS (Custom) & 3,5701 & \textbf{3,5940}\\
IS (Gated) & \textbf{3,6868} & 3,6461\\
\hline
FID (CE) & 14,8877 & \textbf{13,2659}\\
FID (Custom) & 8,4458 & \textbf{3,0337}\\
FID (Gated) & 2,4400 & \textbf{2,1159}\\
\hline
\end{tabular}
\caption{Quantitative comparison between the fixed and the progresive growing mask using the state-of-the-art models and our custom model on the CelebA dataset.}
\label{tab:celeba}
\end{center}
\end{table}

\subsection{Qualitative Evaluation} 
\label{sec:qualitative}

\begin{figure*}[h]
  \centering
  \includegraphics[keepaspectratio, width=0.9\textwidth]{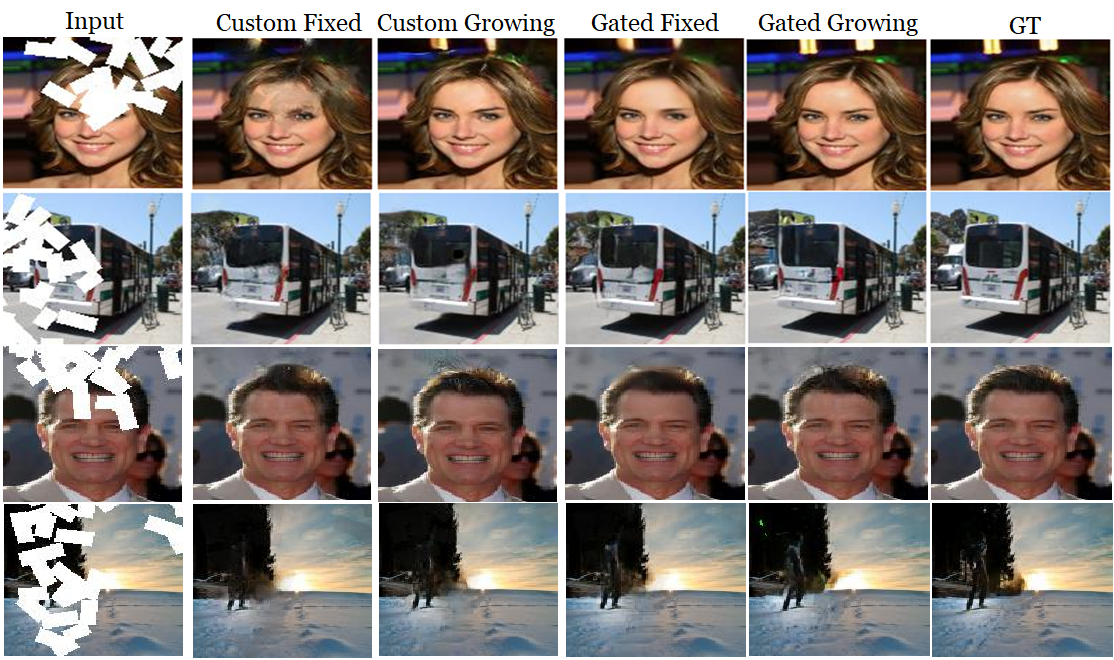}
  \caption{Qualitative results of our Custom and \cite{20} using the fixed and our training approach on the MSCOCO and CelebA datasets.}
  \label{quality_free_form}
\end{figure*}

\begin{figure*}[h]
  \centering
  \includegraphics[keepaspectratio, width=0.6\textwidth]{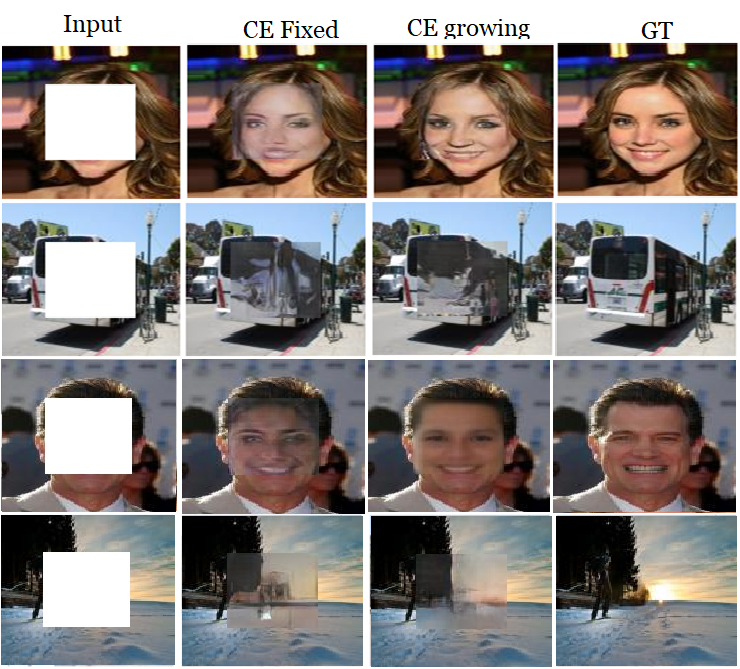}
  \caption{Qualitative results of \cite{17} using the fixed and our training approach on the MSCOCO and CelebA datasets.}
  \label{quality_ce}
\end{figure*}

We compare the fixed and the progressive growing training approach using our customized model and \cite{20} on the MSCOCO and CelebA datasets. Seen from Figure~\ref{quality_free_form}, the custom model does not generate visually realistic images on the fixed setup. Our proposed training approach improves it to complete the missing parts more robustly, but it still generates artifacts compared with \cite{20}. The later can generate smooth and plausible images without our training approach. However, blurriness appears when we increase the mask size. On the other hand, applying the progressive growing approach to \cite{20} composes a stable model that completes the masked parts in a very natural and realistic way and with fewer artifacts. The model in \cite{17} uses a rectangular shape mask in the center of the image. For this reason, we compare only the fixed versus growing the training approaches. Figure~\ref{quality_ce} shows that although applying our approach to that model does not give plausible and natural images, it improves the results of the original model by removing the artifacts around the rectangular mask.

\section{Conclusion}
\label{sec:conclusion}

In this paper, we proposed a new curriculum-style method for image inpainting by progressively growing the mask regions. Experiments show that our method generates realistic and plausible images, even with large masked regions. Further, it improves several inpainting models quantitatively, including the state-of-the-art for a wide variety of regular and irregular masks on MSCOCO and Celeb datasets. In our next work, we aim to combine the progressively growing approach with the proposed training setups to check whether the stability and performances will be improved or not. Furthermore, we want to adopt this curriculum learning approach for other computer vision tasks, including super-resolution and de-blurring.

{\small
\bibliographystyle{unsrt}
\bibliography{egbib}
}

\end{document}